\documentclass[10pt,twocolumn,letterpaper]{article}

\usepackage{cvpr}              

\usepackage{graphicx}
\usepackage{amsmath}
\usepackage{amssymb}
\usepackage{booktabs}
\usepackage{multirow}
\usepackage{amsmath,bm}
\usepackage[accsupp]{axessibility}
\usepackage[pagebackref,breaklinks,colorlinks]{hyperref}

\usepackage[capitalize]{cleveref}
\crefname{section}{Sec.}{Secs.}
\Crefname{section}{Section}{Sections}
\Crefname{table}{Table}{Tables}
\crefname{table}{Tab.}{Tabs.}

\def\cvprPaperID{1842} 
\def\confName{CVPR}
\def\confYear{2023}

\begin{document}

\title{Mobile User Interface Element Detection Via Adaptively Prompt Tuning}


\author{%
Zhangxuan Gu, 
Zhuoer Xu,
Haoxing Chen,
Jun Lan,
Changhua Meng, 
Weiqiang Wang\\
\text{Tiansuan Lab, Ant Group}\\
{\tt\small\{guzhangxuan.gzx,xuzhuoer.xze,chenhaoxing.chx,yelan.lj,changhua.mch,weiqiang.wwq\}}\\
{\tt\small@antgroup.com}
}

\maketitle

\begin{abstract}
  Recent object detection approaches rely on pretrained vision-language models for image-text alignment. However, they fail to detect the Mobile User Interface (MUI) element since it contains additional OCR information, which describes its content and function but is often ignored. In this paper, we develop a new MUI element detection dataset named MUI-zh and propose an Adaptively Prompt Tuning (APT) module to take advantage of discriminating OCR information. APT is a lightweight and effective module to jointly optimize category prompts across different modalities. For every element, APT uniformly encodes its visual features and OCR descriptions to dynamically adjust the representation of frozen category prompts. We evaluate the effectiveness of our plug-and-play APT upon several existing CLIP-based detectors for both standard and open-vocabulary MUI element detection. Extensive experiments show that our method achieves considerable improvements on two datasets. The datasets is available at \url{github.com/antmachineintelligence/MUI-zh}.
\end{abstract}

\section{Introduction}
\label{sec:intro}

While significant progress has been made in object detection\cite{ren2015faster,cai2018cascade,redmon2016you,tian2019fcos,liu2016ssd}, with the development of deep neural networks, less attention has been paid to its challenging variant in the Mobile User Interface (MUI) domain\cite{bunian2021vins}. Instead of personal computers and books, people nowadays spend more time on mobile phones due to the convenience of various apps for daily life. However, there may exist some risks, including illegal gambling\cite{gao2021demystifying,9043521}, malware\cite{wang2020beyond,wang2019rmvdroid}, security\cite{faruki2014android,chen2022illegal}, privacy\cite{lin2012expectation,kim2013detecting}, copy/fake\cite{8804455} and fraudulent behaviors\cite{dong2018frauddroid,hu2020mobile} in apps, which need to be detected and alarmed as required by government authorities and app markets. In apps, these risks may occur in one element or even hide in the subpage after clicking one element. As a result, it is in great need of an accurate, robust, and even open-vocabulary MUI element detection approach in practice.
Such technology can benefit a great variety of scenarios as mentioned above, towards building a better mobile ecosystem\cite{hu2020mobile,viennot2014measurement}.

\begin{figure}[t]
  \centering
  \includegraphics[width=\linewidth]{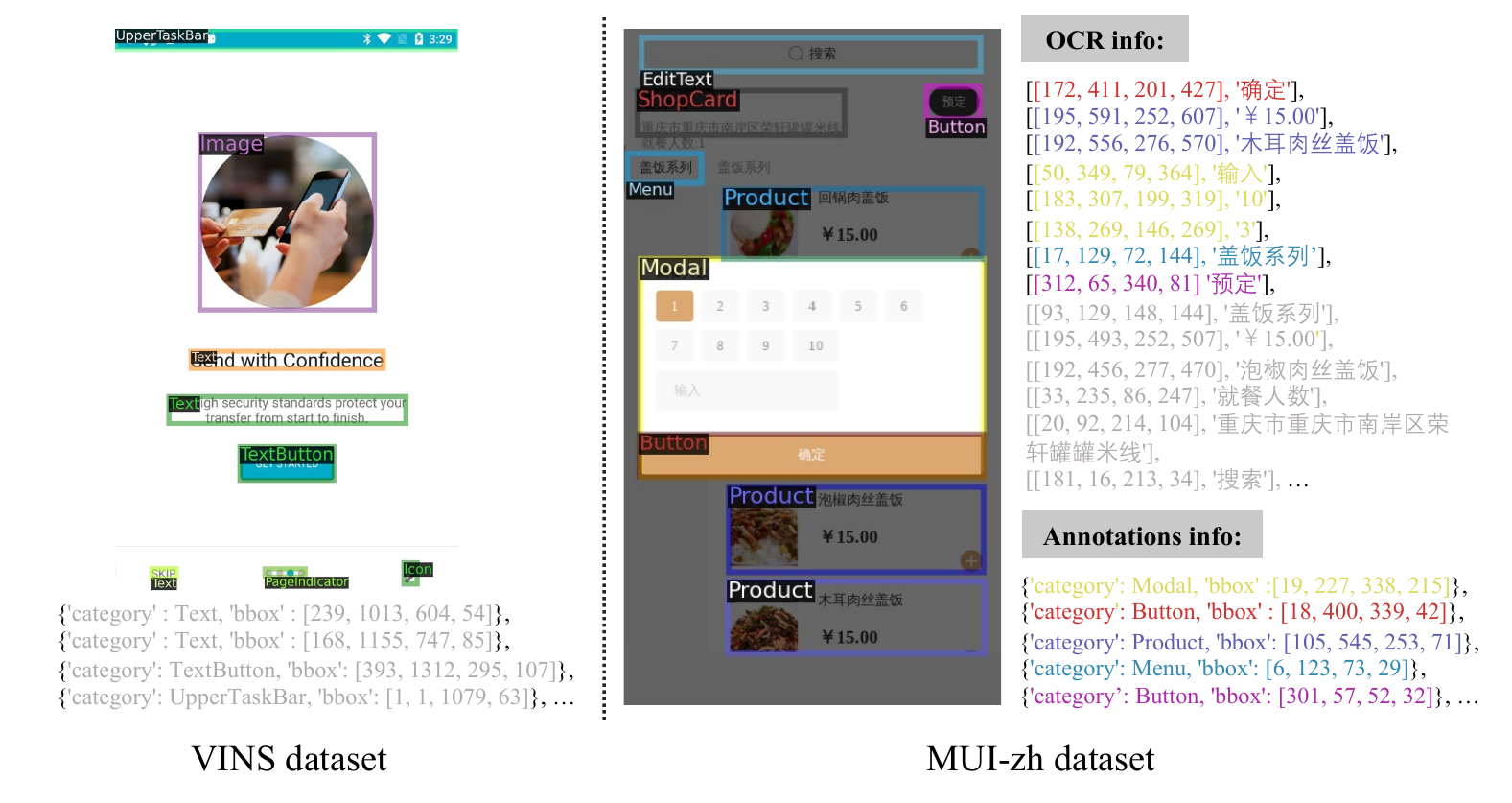}
   \caption{\textbf{Two MUI samples from VINS and MUI-zh dataset.} Compared to VINS, we additionally obtain the OCR descriptions as supplemental information in MUI-zh. Moreover, we further link OCR descriptions and element annotations with the same color.}
   \label{fig:example}
   \vspace{-2px}
\end{figure}

This paper proposes MUI element detection as a variant object detection task and develops a corresponding dataset named MUI-zh. In general, object detection aims to classify and locate each object, such as an animal or a tool, in one raw image. While in MUI data, our primary concern is detecting elements, \emph{e.g.}, products and clickable buttons in the screenshots. 
The main difference between the two tasks is that MUI data often have discriminative OCR descriptions as supplemental information for every element, significantly influencing detection results. To better explain it, we put two MUI data examples from VINS\cite{bunian2021vins} and our MUI-zh in Figure~\ref{fig:example}. VINS only provides the category annotation and bounding box for every element, as the object detection dataset does. At the same time, MUI-zh additionally obtains the OCR descriptions and links them with elements for further usage. Since the OCR descriptions are texts and will be an additional input modality, it is natural to leverage recent Open-Vocabulary object Detection (OVD) models\cite{Hanoona2022Bridging,zhong2022regionclip,zhou2022detecting,zhao2022exploiting,chen2022open,gu2021open,ma2022open} as the MUI element detection baseline because of their rich vision-language knowledge learned from pretrained CLIP\cite{radford2021learning}.


\begin{figure}[t]
  \centering
  \includegraphics[width=\linewidth]{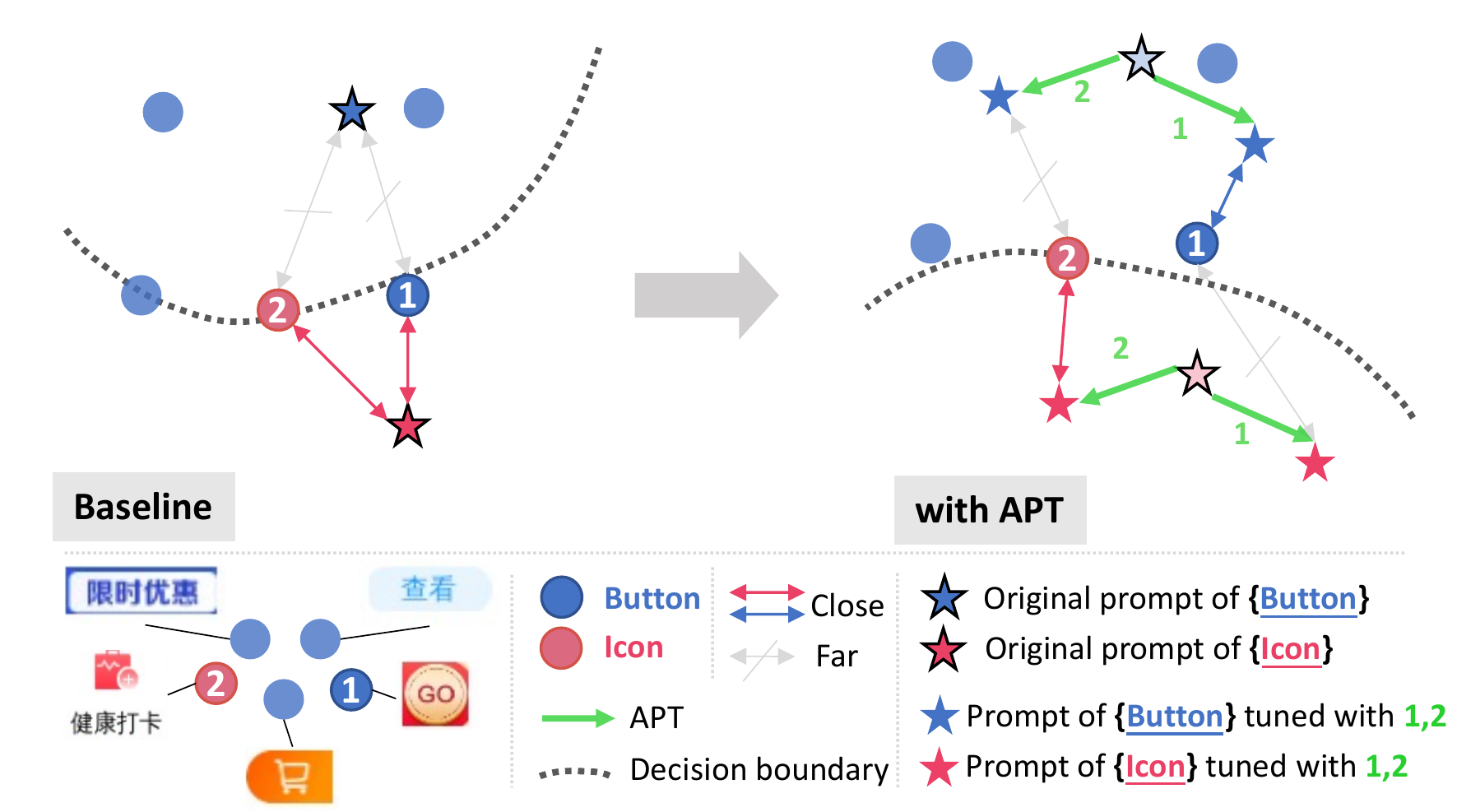}
   \caption{\textbf{Decision boundaries of baseline and adding APT during vision-language alignment.} The stars are category prompts, and the circles are element vision embeddings. Element 1 is misclassified by baseline while our APT tunes its category prompts adaptively and thus successfully matches it and its category.}
   \label{fig:moti}
   \vspace{-2px}
\end{figure}

OVD detectors usually detect and classify objects by calculating the similarity between visual embeddings and textual concepts split from captions. However, according to our experiments, existing OVD methods can not achieve satisfactory performances on MUI datasets. The reason mainly comes from two aspects: Firstly, the samples for training OVD detectors are appearance-centric, while MUI data is not. Besides the appearance, the category of one MUI element is often closely related to its textual explanations obtained by OCR tools. Thus, OCR descriptions of one element can be viewed as a discriminative modality to distinguish itself from other categories, but neither exists nor is used in OVD models; 
Secondly, the category prompts with only category name is not optimal for vision-language alignment since they may not be precise enough to describe an MUI element. For example, we show four buttons (blue) and one icon (red) in Figure~\ref{fig:moti}. The baseline (OVD detector) only uses ``a photo of category name" to perform alignment and misclassify button 1 as an icon.

To alleviate the above issues, we propose a novel lightweight and plug-and-play Adaptively Prompt Tuning (APT) module in MUI element detection. Firstly, it takes OCR descriptions as input, using a unimodal block to obtain rich elements' information (\emph{e.g.}, content and function) for vision-language alignment;
Secondly, it adaptive encodes vision and OCR description features into embeddings to adjust the representation of frozen category prompts, which further reduces the impact of language ambiguity during matching. 
As shown in Figure~\ref{fig:moti}, the gray dotted lines indicate the decision boundaries of the OVD baseline and its variant with APT during the recognizing phase. Element 1 is misclassified by the baseline since its embedding is close to the frozen category prompt of ``icon" and far away from its groundtruth ``button". 
Our APT adaptively tunes two category prompts (noted by the green arrow) for every element and successfully recognizes element 1. As a result, we demonstrate that the APT can achieve noticeable performance gains based on previous OVD detectors, which will benefit many mobile layout analyses\cite{zhang2022effects,zhang2019mobile} and risk hunters\cite{chen2022illegal,gao2021demystifying}. We summarize our contributions as follows.

\begin{itemize}
    \item We develop a high-quality MUI dataset (called MUI-zh) containing 18 common categories with OCR descriptions as the supplemental information. Besides MUI-zh, we will also provide the OCR descriptions of the existing dataset VINS to facilitate future research.
    \item Inspired by the MUI data characteristics, we further proposed a novel Adaptive Prompt Tuning (APT) module to finetune category prompts for standard and open-vocabulary MUI element detection. 
    \item Experiments on two datasets demonstrate that our APT, as a plug-and-play module, achieves competitive improvements upon four recent CLIP-based detectors.
    
\end{itemize}

\section{Related Works}
\label{sec:related}

\subsection{Object Detection}

Object detection aims to detect and represent objects at a bounding box level. There are two kinds of object detection methods, {\em i.e.,} two-stage\cite{ren2015faster,cai2018cascade}, and single-stage\cite{redmon2016you,tian2019fcos,liu2016ssd}. Two-stage methods first detect objects, then crop their region features to further classify them into the foreground or background. In contrast, the one-stage detectors directly predict the category and bounding box at each location.

\subsection{Open-vocabulary Object Detection}

Relying heavily on visual-language pretrained models\cite{radford2021learning}, open-vocabulary object detection approaches aim to locate and classify novel objects that are not included in the training data. Recently, OVD methods\cite{du2022learning,Hanoona2022Bridging,zhong2022regionclip,zhou2022detecting,zhao2022exploiting,chen2022open,gu2021open,ma2022open} follow two-stage fashion: class-agnostic proposals are firstly generated by RPN\cite{ren2015faster} trained on base categories, then the classification head is required to recognize novel classes with the knowledge from pretrained CLIP\cite{radford2021learning}. 


The representative solutions include OVR-CNN\cite{zareian2021open} and ViLD\cite{gu2021open}. Taking Faster RCNN\cite{ren2015faster} as the backbone, OVR-CNN\cite{zareian2021open} trains a projection layer on image-text pairs with contrastive learning, while ViLD\cite{gu2021open} proposes to explicitly distill the knowledge from the pretrained CLIP visual encoder. Advanced to them, Detic\cite{zhou2022detecting} tries to self-train the detector on ImageNet21K\cite{russakovsky2015imagenet} for OVD. Recently, VL-PLM\cite{zhao2022exploiting} use self-training in both two stages on unlabeled data and MEDet\cite{chen2022open} proposes an online proposal mining method to refine the vision-language alignment. Following Detic\cite{zhou2022detecting}, Object-centric OVD\cite{Hanoona2022Bridging} combines knowledge distillation and contrastive learning, achieving the best performance on COCO\cite{lin2014microsoft} with the extra weakly supervised data from ImageNet21K. One closely related work is RegionCLIP\cite{zhong2022regionclip}, which leverages a CLIP model to match image regions with template texts on large-scale data from the web and then uses pseudo pairs to train the fine-grained alignment between image regions and text spans. 

\subsection{Prompts Learning}

The large vision-language model, \emph{e.g.}, CLIP\cite{radford2021learning}, has significantly improved many few-shot or zero-shot computer vision tasks. They are often pretrained on a large amount of image-text pairs collected from the web and can be easily transferred to numerous downstream tasks with either finetuning\cite{lu2019vilbert,su2019vl} or prompt learning\cite{zhou2022learning}. From \cite{radford2021learning}, we can observe that a task-specific prompt can boost performance significantly but needs carefully tuning prompts by humans. As its extension, CoOp\cite{zhou2022learning} proposes context optimization with learnable vectors for automating prompt learning in few-shot classification, relieving the burden of designing hand-craft prompts by humans. Moreover, its further extension CoCoOp\cite{zhou2022conditional} learns a lightweight neural network to generate for each image an input-conditional token, which improves the generalization ability to wider novel categories in image classification tasks.

Recently, DetPro\cite{du2022learning} and PromptDet\cite{feng2022promptdet} adapt CoOp\cite{zhou2022learning} to OVD by designing particular strategies to handle foreground and background proposals within images.
Although the vision embeddings learned in our APT are somehow inspired by CoCoOp\cite{zhou2022conditional}, we are the first to propose a unified module for tuning prompts on two modalities, \emph{i.e.}, OCR descriptions and vision features.

\section{Mobile User Interface Dataset}

\begin{figure*}[ht]
  \centering
  \includegraphics[width=\linewidth]{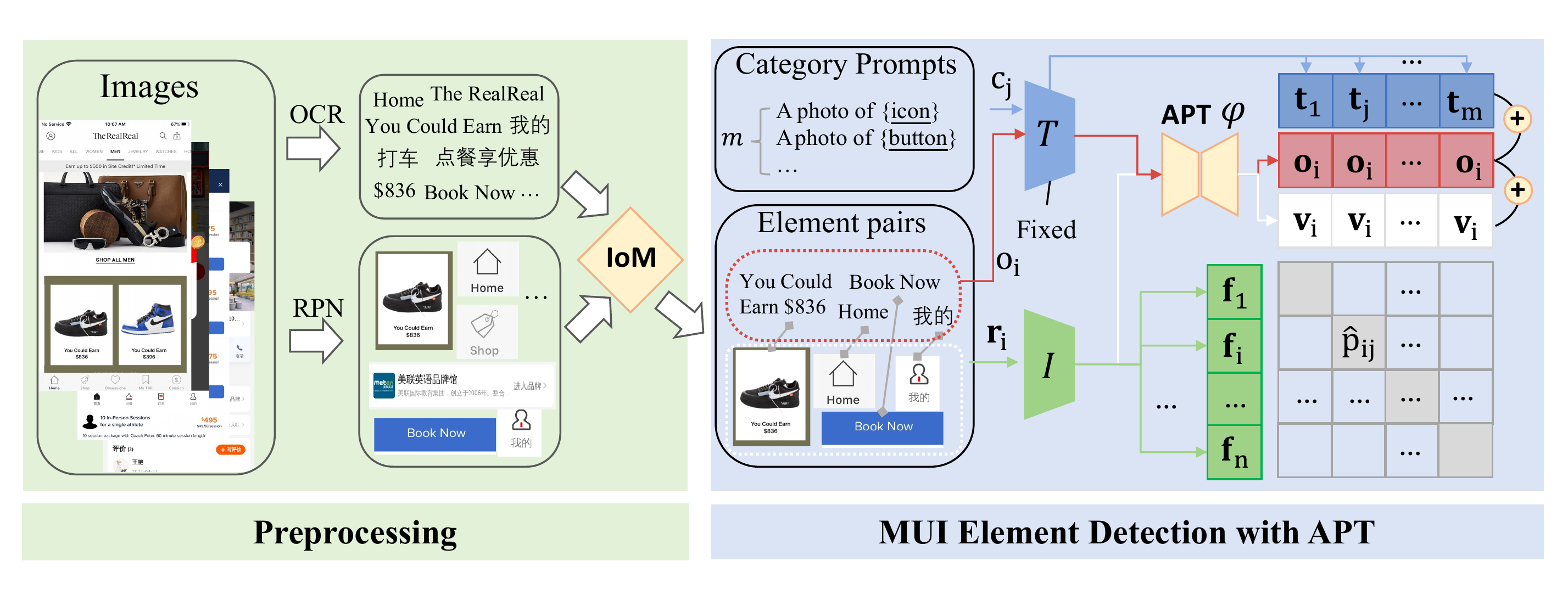}
   \caption{\textbf{Overview of MUI element detection pipeline associated with our proposed APT.} We first use OCR tools and class-agnostic RPN for input images to obtain OCR descriptions and element proposals. An IoM module matches and links the elements and OCR descriptions in the preprocessing phase. Existing CLIP-based detectors usually encode element proposals and category prompts into vision ({\color{green} green}) and text ({\color{blue} blue}) embeddings by image encoder $\bm{I}$ and text encoder $\bm{T}$ for similarity calculation. Our APT additionally uses OCR descriptions ({\color{red} red}) and vision embeddings to tune the text embedding for better alignment. Best viewed in color.}
   \label{overview}
   \vspace{-2px}
\end{figure*}

In this section, we first introduce the existing MUI datasets and our developed MUI-zh. Then we briefly describe how to match OCR descriptions and elements.

\subsection{Dataset Preparation}

Early work on the MUI dataset explored how to support humans in designing applications. For example, Rico\cite{deka2017rico}, a dataset of Android apps, was released five years ago. It consists of 72k MUI examples from 9722 apps, spanning 27 categories in the Google Play Store. However, the annotations of Rico are noisy, sometimes even incorrect, according to \cite{bunian2021vins}. As its extension, VINS\cite{bunian2021vins} uses MUI designs and wireframes to enable element detection and retrieval. 

Nowadays, more and more tinyapps (in apps) are developed by merchants, and their elements, also as MUI data, have a noticeable domain gap with the elements in Rico and VINS. In order to fully understand MUI data, we develop MUI-zh, an MUI detection dataset from tinyapp screenshots. MUI-zh has 5769 images of tinyapp screenshots, including 50k elements within 18 categories. Besides element location and category, we also provide essential OCR descriptions and locations for every screenshot as supplemental information for classification. 
Another reason for developing MUI-zh is that the existing language of MUI datasets is English. Detectors trained on them can not be used in another language, such as Chinese, due to the domain gap/bias during vision-language alignment. Our MUI-zh collects high-quality tinyapp screenshots in Chinese, which enriches the MUI data for different languages.

\subsection{OCR Descriptions Matching}

After we collect and annotate enough MUI screenshots, we have to link the OCR descriptions and elements for further usage. How to relate OCR and elements with their locations is an open question. Intuitively, it is possible to link them by calculating and ranking their Intersection Over Union (IoU) according to two series bounding boxes inspired by non-maximum suppression (NMS). For every element box, we select the OCR boxes whose IoU scores are larger than a threshold (\emph{e.g.}, 0.5) as its descriptions without replacement. Note that OCR tools may separate one sentence into many phrases, and as a result, an element may also be linked to more than one OCR description. Another special case is when an element box does not have any description, we assign it an empty word.

Generally speaking, IoU measures how much two proposals overlap and whether they can be assigned with the same instance in the object detection task. However, MUI elements like products and buttons are more likely to include their OCR descriptions (often occupy only a small region, \emph{e.g.}, $10\%$ of element) within the box. In this case, the IoU (0.1) is smaller than the threshold and this element fails to match its description, which is unacceptable. To tackle this problem, we utilize Intersection Over Minimum (IoM) instead of IoU during OCR matching. IoM replacing the area of union with the area of the minimum box in IoU is suitable for MUI data. For the case mentioned above, the IoM is $1$, which means we successfully link the element and its OCR descriptions. Note that we also conduct OCR matching on VINS and release the results.

\section{Methodology}
\label{sec:method}

In this section, we first briefly present how existing OVD models detect elements of MUI data in Section~\ref{sec:muidet} and then show the architecture of APT and how it works in Section~\ref{sec:apt}. Finally, we claim how to assemble APT on four existing detectors in Section~\ref{sec:assemb}. The whole pipeline of MUI element detection is shown in Figure~\ref{overview}.

\subsection{Detectors on MUI}\label{sec:muidet}

\noindent \textbf{Preprocessing:} Given a batch of MUI data, the training pipeline of recent two-stage CLIP-based detectors follows almost the same scheme (detect-then-classify). They first use a class-agnostic RPN\cite{ren2015faster} to obtain element proposals and perform their innovations and improvements during the classification step. Our APT additionally considers OCR descriptions while aligning the element with categories.

\noindent \textbf{Training:} Existing CLIP-based detectors mainly focus on the training of the classifier. Specifically, they first construct human-made prompts (\emph{e.g.}, ``a photo of category name'') and feed them to the frozen language encoder $\bm{T}(\cdot)$ of pretrained CLIP\cite{radford2021learning} as the text embeddings. At the same time, a trainable CLIP visual encoder $\bm{I}(\cdot)$ is adapted to the detector for encoding element proposals into vision embeddings. Finally, the classifier learns to match these pair-wise embeddings via contrastive learning and cross-entropy loss. 

Specifically, assuming an image $I$ has $n$ element proposals obtained by RPN, and we notate their features as $\{\mathbf{r}_i\}_{i=1}^n \in \mathcal{R}^d$. The classifier's goal is to match the element proposals with category prompts $\{{c}_j\}_{j=1}^m$ for $m$ different categories. Relying on the powerful CLIP, the text (vision) embedding $\mathbf{t}_j$ ($\mathbf{f}_i$) of category $j$ (proposal $i$) is generated by feeding $c_j$ ($\mathbf{r}_i$) into the encoders, respectively:
\begin{equation}
    \mathbf{t}_j = \bm{T}({c}_j);
    \mathbf{f}_i = \bm{I}(\mathbf{r}_i).
\end{equation}
For a paired proposal $i$ and its groundtruth category $j$ during training, we can calculate the predicted probability as:
\begin{equation}\label{equ1}
    p_{ij} = \frac{\exp(\cos(\mathbf{t}_j,\mathbf{f}_i)/\tau)}{\sum_{k=1}^m\exp(\cos(\mathbf{t}_k,\mathbf{f}_i)/\tau)},
\end{equation}
where $\tau$ is a temperature hyper-parameter. Finally, the cross-entropy loss is applied to optimize the network parameters except for $\bm{T}$ on proposal $i$:
\begin{equation}
    L_i = -\log(p_{ij}).
\end{equation}
The reason for freezing $\bm{T}$ is to fully utilize the knowledge learned by the CLIP pretrained on large-scale data according to \cite{du2022learning,Hanoona2022Bridging}. We also conduct experiments to verify it. 

\noindent \textbf{Inference:} OVD models predict the category with the probability obtained by Equation~\ref{equ1} for the element detection task. While performing experiments in the open-vocabulary setting, we extend the category prompts to cover both base and novel classes following \cite{zhong2022regionclip}.

\subsection{Adaptively Prompt Tuning}\label{sec:apt}

As we mentioned in Section~\ref{sec:intro}, existing CLIP-based detectors are not generalizable to MUI categories due to the ignorance of OCR descriptions and the difficulty of aligning various-appearance elements to one frozen manual category prompts. To deal with these two weaknesses, we propose an Adaptively Prompt Tuning (APT) module by mapping OCR descriptions (red) and vision embeddings (green) into the space of text embeddings to adaptively tune the category prompts for every element proposal as shown in Figure~\ref{overview}. The figure shows that the mapped embeddings (red and white) are fused to adjust the frozen text embeddings (blue) for final alignment with vision embeddings (green).

For simplicity, we use $\varphi(\cdot)$ to denote the APT and formulate the training pipeline for image $I$ as:
\begin{equation}\label{equ3}
    \mathbf{o}_i = \varphi(\bm{T}({o}_i));
    \mathbf{v}_i = \varphi(\mathbf{f}_i);
    \mathbf{\hat{t}}_{ji} = \mathbf{t}_j+\mathbf{o}_i+\mathbf{v}_i;
\end{equation}

\begin{equation}\label{equ2}
    \hat{p}_{ij} = \frac{\exp(\cos(\mathbf{\hat{t}}_{ji},\mathbf{f}_i)/\tau)}{\sum_{k=1}^m\exp(\cos(\mathbf{\hat{t}}_{ki},\mathbf{f}_i)/\tau)},
\end{equation}
where $\{o_i\}_{i=1}^n$ are the OCR descriptions for $n$ proposals. In this way, we can optimize the whole model except for $\bm{T}$ with cross-entropy loss:
\begin{equation}
    \hat{L}_i = -\log(\hat{p}_{ij}).
\end{equation}
Note that during inference, we also tune the text embeddings in the same way as training with Equation~\ref{equ3}.

Since our goal is to map supplemental information into the embedding space for prompt tuning, it is natural to uniformly encode OCR descriptions and vision embeddings to encourage knowledge sharing and interaction from different modalities. As we know, APT is the first unimodal prompt tuning method, holding higher performances than individually encoding two modalities with different network parameters, as shown in our experiments.

Inspired by CoCoOp\cite{zhou2022conditional}, we construct APT as a lightweight network with only two bottlenecks, which contains a fully-connected layer (fc) associated with a batch norm (bn) and a relu activation. It follows standard encoder-decoder fashion, and the fc is utilized to reduce/enlarge the number of feature channels (16x). Since the input channel of the visual feature is 1024, the total number of parameters of APT is about 128k, including the weights and bias, which have little influence on training and inference speed.

At the end of APT, we also explore how to fuse modality information in three ways: element-wise sum, element-wise multiply, and fusion with fc. Recall that the attention mechanism\cite{vaswani2017attention} is also influential in modality fusion and feature extraction. When we choose element-wise sum as the fusion function, our APT works as an attention layer for different modalities except for the self-attention part calculated on $\mathbf{t}_j$ in equation~\ref{equ3}, which is a constant. If we use fc to learn the weights for fusion, then $\mathbf{t}_j$ can also be learned, which means our APT, in this case, has the same function of attention layers. According to our experiments, we use element-wise sum as the fusion function due to the slightly higher performance and lower calculating complexity.

In conclusion, we highlight that our goal of APT is \emph{to adaptively tune frozen category prompts with the context from every element's OCR description and specific vision information}. 
Another interesting thing is that there exist many variants of APT. For example, what if we tune the category prompts only with vision embeddings and tune vision embeddings with OCR descriptions? Moreover, can we tune vision embeddings by self-attention and OCR descriptions while leaving category prompts fixed? 
To explore the influence of different tuning methods mentioned above, we conduct experiments in Section~\ref{sec:ablation}.

\subsection{Assembling APT to CLIP-based Detectors}\label{sec:assemb}

DetPro\cite{du2022learning}, PromptDet\cite{feng2022promptdet}, Object-centric\cite{Hanoona2022Bridging}, and RegionCLIP\cite{zhong2022regionclip} are recent CLIP-based frameworks for OVD. As we mentioned, our APT tunes category prompts without changing model architectures and thus can be used directly by many OVD methods. Here we explain how and where to equip them with APT in detail. Firstly, RegionCLIP\cite{zhong2022regionclip} and Object-centric\cite{Hanoona2022Bridging} use the fixed manual prompts, and we can easily add APT upon them at the end of the network during classification. For PromptDet\cite{feng2022promptdet} and DetPro\cite{du2022learning}, they both use the CoOp\cite{zhou2022learning} to generate trainable category prompts instead of manual ones. Our APT adjusts that trainable category prompts for fair comparisons in this case. 




\section{Experiments}
\label{sec:exp}

In this section, we first introduce the implementation details for datasets and models in Section~\ref{sec:detail}. Our main results are APT upon CLIP-based detectors for both standard and open-vocabulary MUI element detection as shown in Section~\ref{sec:sota}. Moreover, we evaluate the ablations to study model components in Section~\ref{sec:ablation}. Since the bounding boxes annotated by Rico\cite{deka2017rico} are noisy according to \cite{bunian2021vins}, we only conduct experiments on MUI-zh and VINS for comparison. Finally, we evaluate our APT for the object detection task on COCO\cite{lin2014microsoft} in Section~\ref{sec:coco}.

\subsection{Implementation Details}\label{sec:detail}

\noindent \textbf{Datasets.} We evaluate our method on two MUI element detection datasets, namely MUI-zh and VINS\cite{bunian2021vins}. MUI-zh is a high-quality MUI element detection dataset with screenshots collected from mobile tinyapps. Its training set contains 4769 images and 41k elements, while the validation set has 1000 images and 9k elements within 18 categories. Another popular MUI dataset is VINS\cite{bunian2021vins}, which contains 3826 training and 981 validation images with 20 categories. For open-vocabulary element detection, we set the product, icon, button, card, tips, and menu as the base categories and the remaining 12 elements as novel ones on MUI-zh. As for VINS, we set background-image, card, text and spinner as four novel categories and others as base categories.

\noindent \textbf{Training details and metrics.} We evaluate MUI element detection performance on MUI-zh and VINS for both standard and open-vocabulary settings. During training, the default visual encoder of all models we used in the experiments is ResNet50\cite{he2016deep} from pretrained CLIP\cite{radford2021learning}. Note that the language encoder is frozen following \cite{du2022learning,Hanoona2022Bridging}. For MUI element detection, SGD is used with a batch size of 64, an initial learning rate of 0.002, and a maximum iteration of 12 epochs on 8 A100 GPUs.
For open-vocabulary element detection, RPN is trained with the base categories of two datasets. The temperature $\tau$ is 0.01.
The widely-used object detection metrics, including Mean Average Precision (mAP) for novel and all categories are used.

\subsection{Main Results of MUI Element Detection}\label{sec:sota}

\noindent \textbf{MUI element detection.} As shown in Table~\ref{tab:sota}, we list two groups of detection approaches on both MUI-zh and VINS. The first group is standard object detection methods like Faster RCNN\cite{ren2015faster} and Cascaded RCNN\cite{cai2018cascade}, while the second group contains four CLIP-based models.

The table shows that recently proposed Object-centric OVD\cite{Hanoona2022Bridging} and RegionCLIP\cite{zhong2022regionclip} achieve much better performances than standard object detection models since MUI data need more attention on vision-language alignment. Moreover, our APT improves about 4-5\% (6-9\%) mAP on CLIP-based detectors on MUI-zh (VINS), which is a significant enhancement and shows APT's effectiveness in the MUI element detection task. Among these detectors, RegionCLIP\cite{zhong2022regionclip} equipped with APT achieves the best performances (51.23\% and 80.84\% on MUI-zh and VINS).

\begin{table}[t]
\resizebox{0.48\textwidth}{!}{ 
\begin{tabular}{c|c|cc}
\toprule
Methods                     & Publication & MUI-zh & VINS \\  \hline
VINS\cite{bunian2021vins}                 &    CHI'21    & -         & 63.21  \\
Faster RCNN\cite{ren2015faster}                & NeurIPS'15    & 44.63     & 68.89  \\
Cascaded RCNN\cite{cai2018cascade}                & \multirow{2}{*}{CVPR'18}    & 46.76     & 72.85  \\ 
+ COCO pretrain &   & 48.80     & 75.77  \\ \hline
DetPro\cite{du2022learning} & CVPR'22    &  44.55   & 71.67  \\ 
\cite{du2022learning}+APT               & -    & 48.62({\color{red}+4.07})     & 77.73({\color{red}+6.06})  \\ 
PromptDet\cite{feng2022promptdet} &ECCV'22     &  40.14   &  68.94 \\
\cite{feng2022promptdet}+APT                  & -    & 45.07({\color{red}+4.93})    &  76.43({\color{red}+7.49}) \\
Object-centric\cite{Hanoona2022Bridging} &NeurIPS'22     & 45.87    & 72.36  \\
\cite{Hanoona2022Bridging} +APT &-     &  50.78({\color{red}+4.91})   & 79.48({\color{red}+7.12})  \\
RegionCLIP\cite{zhong2022regionclip}        & CVPR'22    & 45.51     & 71.53  \\
\cite{zhong2022regionclip}+APT                  & -    & \textbf{51.23}(\textbf{{\color{red}+5.72}})     & \textbf{80.84}(\textbf{{\color{red}+9.31}})  \\ 
\bottomrule
\end{tabular}}
\caption{\textbf{Results (mAP\%) of MUI element detection.} We list the performance of six popular object detection approaches based on ResNet50. Besides them, we additionally report the performance gains of our APT module over four recent CLIP-based models.}
\label{tab:sota}
\vspace{-2px}
\end{table}

\noindent \textbf{Open-vocabulary MUI element detection.} One more advantage of CLIP-based detectors compared to object detection ones is that they can detect objects not in the predefined categories. To this end, we also conduct experiments on open-vocabulary MUI element detection, and the results are in Table~\ref{tab:openset}. Here we compare four recent methods with and without our APT on two datasets. The table shows that APT achieves noticeable improvements upon the listed methods. More specifically, among four CLIP-based methods, Object-centric OVD\cite{Hanoona2022Bridging} with APT outperforms others on the MUI-zh, while RegionCLIP associated with APT gets the best performance on VINS.

Note that even though we have 80\% (16/20) base categories on VINS, the performances of these methods on novel categories still need improvement compared to OVD detectors on COCO novel categories. There are mainly two reasons. Firstly, compared to COCO, the category names of MUI data have much less relation, which causes difficulty for knowledge transfer and embedding alignment. For example, the knowledge of recognizing cats can be easily transferred to classify dogs, while it is challenging to utilize the knowledge of recognizing cards for classifying icons in MUI data. For example, Drawer and Switch are two MUI categories in VINS. However, they have at least two meanings (polysemy words) with various appearances (real-world object and MUI element), making the transfer difficult. Another limitation is the size of the dataset. For open-vocabulary detection on COCO, there are thousands of ($>$110k) annotated object-caption pairs, not to mention numerous unpaired data from the web for self-supervised training. At the same time, the recent MUI datasets only have less than 10k samples for training.

\begin{table}[t]
\resizebox{0.48\textwidth}{!}{ 
\begin{tabular}{c|cc|cc}
\toprule
\multirow{2}{*}{Methods}   & \multicolumn{2}{c|}{MUI-zh} & \multicolumn{2}{c}{VINS} \\ 
                           & Novel(12)            & All(18)             & Novel(4)             & All(20)           \\ \hline
DetPro\cite{du2022learning}     &  0.54   & 15.99 &   2.03  & 54.68\\
\cite{du2022learning}+APT     & 1.41 ({\color{red}+0.87})   & 16.93 ({\color{red}+0.94})&  2.74(\textbf{{\color{red}+0.71}})   & 54.95({\color{red}+0.27})\\
PromptDet\cite{feng2022promptdet}      & 0.78    & 17.02 & 2.59    & 54.86\\
\cite{feng2022promptdet} +APT     & 1.83 ({\color{red}+1.05})   & 18.12(\textbf{{\color{red}+1.10}}) &  3.02 ({\color{red}+0.43})  & 55.18({\color{red}+0.32})\\
Object-centric\cite{Hanoona2022Bridging}  &  1.31   & 17.28 &    3.19 & 55.34\\
\cite{Hanoona2022Bridging}+APT  &  \textbf{2.36}(\textbf{{\color{red}+1.05}})   & \textbf{18.36}({\color{red}+1.08})  & 3.76 ({\color{red}+0.57})   &55.49({\color{red}+0.15}) \\
RegionCLIP\cite{zhong2022regionclip}            &   1.06  & 17.34 & 3.61    &55.79 \\
\cite{zhong2022regionclip} +APT              &  2.10 ({\color{red}+1.04})                      & 18.23  ({\color{red}+0.89})             &       \textbf{4.23}({\color{red}+0.62})                &     \textbf{56.80({\color{red}+1.01})}                 \\
\bottomrule
\end{tabular}}
\caption{\textbf{Results (mAP\%) of open-vocabulary MUI element detection.} We report the performances of four CLIP-based methods on two datasets. Note that the number of novel categories of MUI-zh is 12, while VINS has four novel classes. Our APT improves the results of both novel and all categories.}
\label{tab:openset}
\end{table}

\subsection{Ablation Studies}\label{sec:ablation}

\begin{table}[t]
\resizebox{0.48\textwidth}{!}{ 
\begin{tabular}{c|c|cc}
\toprule
\multicolumn{2}{c|}{Various APT architectures}           & MUI-zh & VINS \\ \hline
\multirow{3}{*}{Ablation}  & 
 APT($\mathbf{v}_i+\mathbf{o}_i$)      &  \textbf{51.23}       &  \textbf{80.84}  \\ 
                                &  w/o $\mathbf{o}_i$         & 47.96 ({\color{blue}-3.27})  & 75.97({\color{blue}-4.87})  \\
&  w/o   $\mathbf{v}_i$       & 48.91 ({\color{blue}-2.32})    & 76.32({\color{blue}-4.52})  \\ \hline
\multirow{2}{*}{Weights}       & Share   weights    &    \textbf{51.23}     &    \textbf{80.84}  \\
                                   & Individual weights & 51.01({\color{blue}-0.22})       &  79.65 ({\color{blue}-1.19})   \\ \hline
\multirow{2}{*}{Layers}        & 2 (fc+bn+relu)    &\textbf{51.23}     &    \textbf{80.84}      \\
                                   & 3 (fc+bn+relu)    &   51.19 ({\color{blue}-0.04})     &    80.79 ({\color{blue}-0.05}) \\ \hline
\multirow{4}{*}{Tuning} & $\mathbf{t}_j+\mathbf{v}_i+\mathbf{o}_i$   vs. $\mathbf{f}_i$  &  \textbf{51.23}       &  \textbf{80.84}    \\
                                   & $\mathbf{t}_j+\mathbf{o}_i$   vs. $\mathbf{f}_i+\mathbf{v}_i$  &  51.08({\color{blue}-0.15})       & 78.23 ({\color{blue}-2.61})    \\ 
                                    & $\mathbf{t}_j+\mathbf{v}_i$   vs. $\mathbf{f}_i+\mathbf{o}_i$  &  50.23({\color{blue}-0.10})       &  78.35({\color{blue}-2.49})    \\ 
                                   & $\mathbf{t}_j$ vs. $\mathbf{f}_i+\mathbf{v}_i+\mathbf{o}_i$    &    51.00 ({\color{blue}-0.23})    &  77.97 ({\color{blue}-2.87})   \\ \hline
\multirow{3}{*}{Fusion}        & Element-wise sum        &       \textbf{51.23}       &  \textbf{80.84}     \\
                                   & Element-wise multi       &   48.83  ({\color{blue}-2.40})    &  77.65  ({\color{blue}-3.19})  \\
                                   & Attention(Concat + fc)            &  51.17 ({\color{blue}-0.06})      &  80.59 ({\color{blue}-0.25})\\  \hline
\multirow{2}{*}{Encoder}        & Freeze $\bm{T}$   &       \textbf{51.23}       &  \textbf{80.84}     \\
                                   & Trainable $\bm{T}$       &     45.11({\color{blue}-6.12})    & 73.13({\color{blue}-7.71})  \\
\bottomrule
\end{tabular}
}
\caption{\textbf{Ablation studies of APT and its variants.} We evaluate APT variants from several views, including their architecture, ablation and tuning methods.} 
\label{tab:ablation}
\vspace{-5px}
\end{table}


We perform experiments for the ablation studies on two datasets. First, we show the impact of progressively integrating our two tuning modalities: the OCR descriptions $\mathbf{o}_i$ and vision embeddings $\mathbf{v}_i$, to the baseline RegionCLIP in Table~\ref{tab:ablation}. Then we explore different settings of weights, layers, tuning methods, and fusion functions, respectively.

\begin{figure*}[ht]
  \centering
  \includegraphics[width=\linewidth]{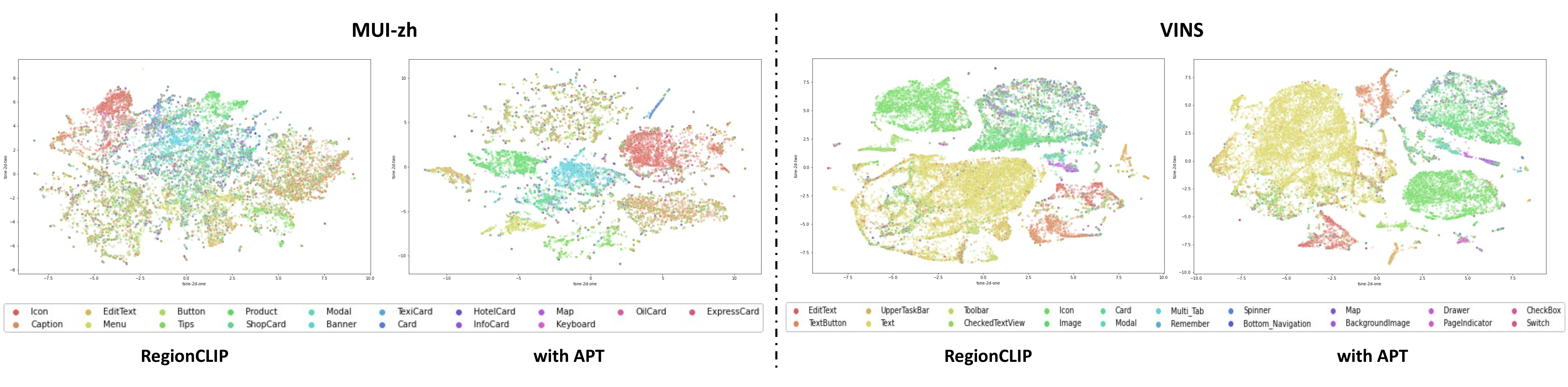}
   \caption{\textbf{T-SNE visualizations.} We perform RegionCLIP without and with APT on two datasets. T-SNE is utilized to visualize their region embeddings. It shows that APT contributes a lot to vision-language alignment. Best viewed in color and in-zoom.}
   \label{fig:tsne}
   \vspace{-2px}
\end{figure*}

\noindent \textbf{Analysis for Components.} As shown in Table~\ref{tab:ablation}, we first use only the vision embeddings $\mathbf{v}_i$ to tune the category prompts, which decreases about 3.3\% (4.9\%) mAP on MUI-zh (VINS). It means that the OCR descriptions of one element contribute a lot to its classification result. In the next row, we only equip RegionCLIP with OCR descriptions $\mathbf{o}_i$. Removing $\mathbf{v}_i$ leads to a 2.3\% (4.5\%) decrease in mAP for two datasets, which means adaptively tuning prompts according to the appearance is also crucial to the final performance. Overall, the whole improvements of APT upon baseline RegionCLIP indicate its effectiveness.

\begin{table}[t]
\resizebox{0.48\textwidth}{!}{ 
\begin{tabular}{c|cc|ccc}
\toprule
\multirow{2}{*}{Method} &   Novel & Base    & \multicolumn{3}{c}{Generalized(17+48)} \\
    &(17) & (48)    & Novel& Base & All \\ \hline
RegionCLIP\cite{zhong2022regionclip}  &35.2 &  57.6   & 31.4& 57.1 & 50.4 \\
\cite{zhong2022regionclip} +APT($\mathbf{v}_i$)  &\textbf{36.3} &  57.3  &\textbf{32.1} & 57.2 &50.0  \\
\cite{zhong2022regionclip} +APT($\mathbf{v}_i+\mathbf{o}_i$)   & {35.9} &  \textbf{57.7}   & {31.8}& \textbf{57.3}& \textbf{50.6}\\
\bottomrule
\end{tabular}
}
\caption{\textbf{Results (mAP\%) of OVD on COCO dataset.} We evaluate APT upon RegionCLIP (backbone ResNet50) following the standard base/novel split setting for a fair comparison.} 
\label{tab:coco}
\vspace{-2px}
\end{table}

\noindent \textbf{Analysis for weights sharing.} Our APT is suitable for two different modalities, as verified in Table~\ref{tab:ablation}. We can observe that using a unified network for encoding OCR and vision embeddings is slightly better than two individual ones with the same architecture. The reason may be that OCR descriptions of one element often describe its appearance. Thus, a lightweight unified network can naturally map two modalities into one semantic space for prompt tuning.

\begin{figure*}[t]
  \centering
  \includegraphics[width=0.84\linewidth]{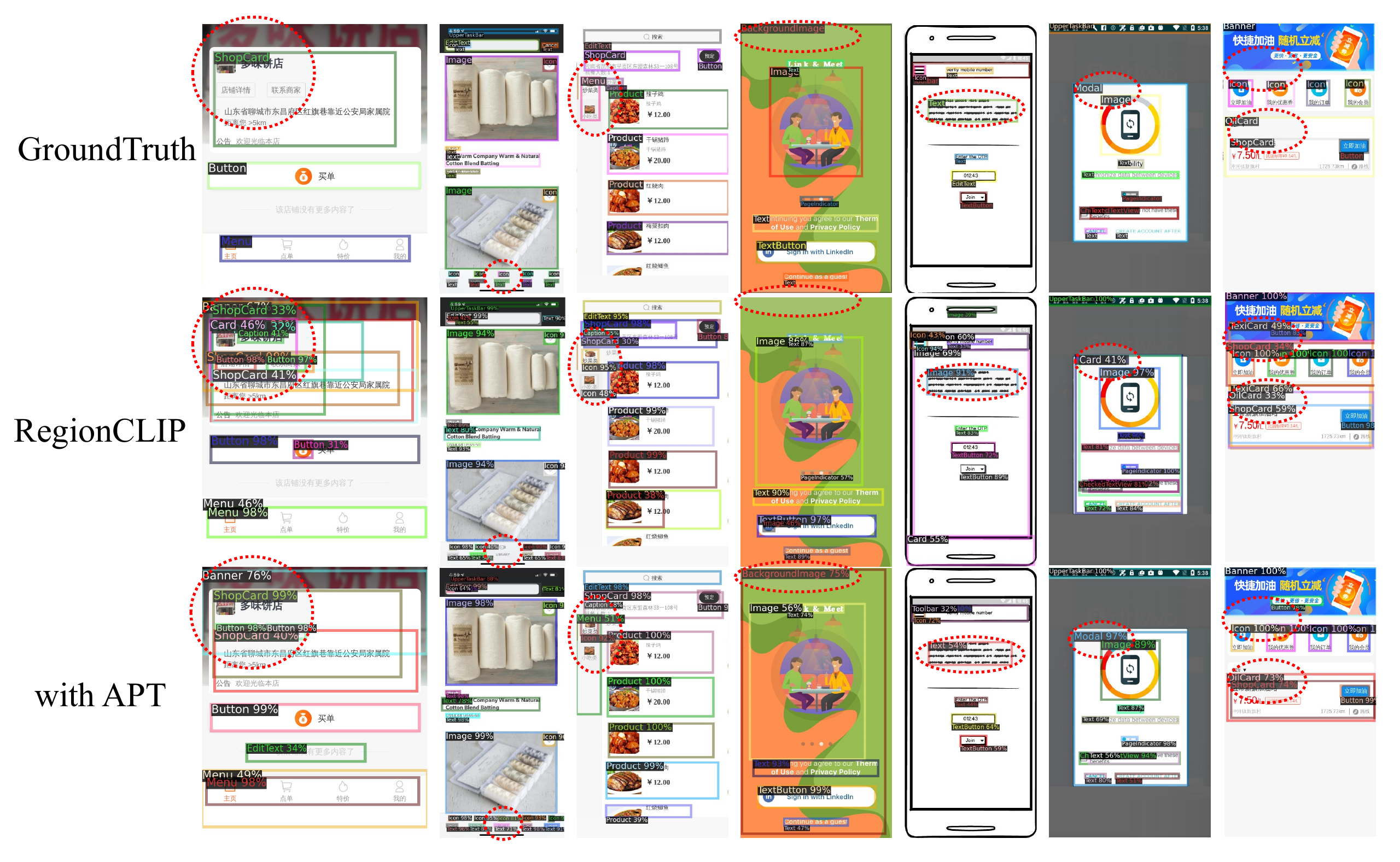}
   \caption{\textbf{Visualizations of MUI element detection.} We successively visualize the images and element bounding boxes of ground truth, RegionCLIP and with our APT. Note that we highlight the differences with the red dotted circle. Best viewed in color.}
   \label{fig:vis}
   \vspace{-2px}
\end{figure*}

\noindent \textbf{Analysis for layers.} Besides the weights of APT, we also want to explore its layer numbers. We compare two settings: 2 or 3 bottlenecks (fc+bn+relu) as presented in Table~\ref{tab:ablation}. With one extra layer, the network performance decreases a little. So our APT can be lightweight (only two layers with 128k parameters) and not time-consuming.

\noindent \textbf{Analysis for tuning methods.} An important part of APT is how to tune the prompts. Since our objective function is the similarity between category prompts and vision embeddings, there are four main ways: tuning only category prompts, tuning only vision embeddings and tuning both as shown in Table~\ref{tab:ablation}. We choose only to tune the category prompts with both OCR and vision embeddings, which gets the best performance. We believe the frozen category prompts rather than the trainable vision embeddings should be tuned adaptively in the MUI data domain.

\noindent \textbf{Analysis for fusion functions.} How to fuse the embeddings from different modalities also impacts element detection results. We compare element-wise sum, multiply and concentration with fc. Among them, the element-wise sum obtains the best performance with no extra parameters. We also believe employing element-wise sum for embedding fusion makes our APT work like an attention layer, excluding the self-attention part calculated on fixed text embeddings.

\noindent \textbf{Analysis for text encoder.} We also show the results of whether to freeze the text encoder $\bm{T}$ in this table. While we train $\bm{T}$ with MUI data, a large performance drop appears. As a result, we follow \cite{du2022learning,Hanoona2022Bridging} to freeze $\bm{T}$ in this paper.

\subsection{Generalization on Object Detection}\label{sec:coco}

Although our APT is specially designed for MUI element detection with extra OCR information, it can also be modified to tune the category prompts on object detection tasks. To this end, we additionally conduct OVD experiments on COCO\cite{lin2014microsoft}.
We follow the data split of \cite{zhong2022regionclip} with 48 base categories and 17 novel categories and we also use the processed data from \cite{zhong2022regionclip} with 110k training images and 4836 test images. Since objects in COCO usually have no OCR descriptions, we directly use their category names as the OCR descriptions, and thus we can build APT on RegionCLIP for the OVD task. 

As shown in Table~\ref{tab:coco}, our APT slightly outperforms RegionCLIP on all metrics (\emph{e.g.}, 31.7 vs. 31.4 on novel categories) in the generalized setting. Compared with RegionCLIP in the standard OVD setting, our APT improves novel categories by about 0.7 mAP but only helps a little on the base categories. We find that the improvements in novel categories are larger than base ones, which indicates the effectiveness of APT in knowledge transfer. With these studies, we conclude that our APT positively impacts MUI element detection and object detection tasks.

\subsection{Visualizations}

\noindent \textbf{T-SNE plots of region vision embeddings.} We have shown that APT can significantly improve performance over the baseline RegionCLIP. However, because the CLIP-based models implicitly learn the alignments by calculating similarity, it is interesting to see their region vision embeddings after training. We show the t-SNE plots of RegionCLIP and APT region embeddings (after non-linear dimensionality reduction) of MUI categories on two validation datasets in Figure~\ref{fig:tsne}. 
We can observe that APT promotes intra-class compactness and inter-class separation, which benefits the vision-language alignment. For example, in MUI-zh, our APT separates products and banners better than RegionCLIP. As for VINS, our model can successfully classify edittexts and textbuttons, while RegionCLIP can not.

\noindent \textbf{Detection on MUI data.} The detection visualizations of RegionCLIP and our APT on two MUI datasets are shown in Figure~\ref{fig:vis}. We successively visualize the images, ground truth element boxes, RegionCLIP and ours. The red dotted circles in this figure highlight the differences. For example, RegionCLIP misclassifies texts in the fifth column and modal in the sixth column, while ours does not. It shows that our APT can better detect elements in MUI datasets.


\section{Conclusion}
\label{sec:conclusion}

In this work, we introduced APT, a lightweight and effective prompts tuning module for MUI element detection. Our APT contains two modality inputs, \emph{i.e.}, element-wise OCR descriptions and visual features. They are fused and encoded within the APT to obtain embeddings for category prompt tuning. It significantly improves performance on existing CLIP-based models and achieves competitive results on two MUI datasets. We also released MUI-zh, a new MUI dataset with matched OCR descriptions. In summary, our model and dataset can benefit various real-world domains, such as robot interaction, information retrieval, targeted advertising, and attribute extraction on mobile phones. We hope our work could inspire designing new frameworks to tackle the challenging MUI element detection tasks. 

\noindent \textbf{Limitations.} Our work has several limitations that can be further investigated. 
(1) The open-vocabulary capabilities of existing models on the MUI data could be further improved compared to the results on OVD datasets as mentioned in Section~\ref{sec:sota}. 
(2) Existing methods all rely on the frozen language encoder from CLIP. We believe the performance drop of unfreezing the language encoder may be due to the small dataset size. 

\section{Acknowledgements}

We thank the contribution of Qiangqiangzhu and his team Tinyapp Ecological Business Team in Ant Group for establishing MUI-zh dataset. Besides, we also thank the Business Risk Management—eKYB Team from the Ant Group for their help in this paper.

{\small
\bibliographystyle{ieee_fullname}
\bibliography{egbib}
}

\end{document}